\documentclass[12pt]{article}

\usepackage{sbc-template}
\usepackage{graphicx,url}
\usepackage[utf8]{inputenc}
\usepackage{subfigure}
\usepackage{xcolor}
\usepackage{adjustbox}
\usepackage{hyperref}

\usepackage{algorithm}

\usepackage{algpseudocode}
\usepackage{amsmath}

\usepackage{amssymb, algorithmicx}
\usepackage{placeins}

\title{Exploring multimodal implicit behavior learning for vehicle navigation in simulated cities}

\author{Eric Aislan Antonelo\inst{1}, Gustavo Claudio Karl Couto\inst{1},Christian Möller\inst{2}}

\address{ Automation and Systems Engineering Department, \\ Federal University of Santa Catarina, Florianopolis, Brazil
  \nextinstitute   
  Faculty of Science and Engineering, Information Technology\\
 Åbo Akademi University, Finland 
  \email{eric.antonelo@ufsc.br, 	gustavo.karl.couto@posgrad.ufsc.br
   }
  \email{
   christian.moller@abo.fi}
}

\begin{document} 

\maketitle

\begin{abstract}
%   Standard Behavior Cloning (BC) often fails to capture the multimodality of real-world driving decisions, where multiple valid actions may exist for the same scenario. 
% In this work, we explore implicit behavioral cloning (IBC) for multimodal imitation learning in autonomous driving, leveraging Energy-Based Models (EBMs) to learn an energy landscape over the action space, instead of mapping an observation directly to the desired action as in BC.
% We introduce a novel Data-Augmented IBC (DA-IBC) method that improves upon conventional IBC by sampling counterexamples from expert data and perturbing them to improve learning efficiency.
% %better represent challenging alternatives. 
% Additionally, we adapt a derivative-free optimization (DFO) procedure with better starting points to improve inference efficiency. 
% Experiments in the CARLA simulator, using a Bird's-Eye View of the vehicle as input, demonstrate that DA-IBC outperforms both BC and standard IBC in routeless urban driving tasks designed to evaluate multimodal behavior learning in a test environment. 
% % by enabling multimodal decision-making at T-intersections of a new test city environment. 
% Visualizations of the learned energy landscapes for DA-IBC and of the agent trajectories confirm the agent’s ability to learn multiple plausible driving strategies in ambiguous situations, such as intersections without explicit route information.
Standard Behavior Cloning (BC) fails to learn multimodal driving decisions, where multiple valid actions exist for the same scenario. We explore Implicit Behavioral Cloning (IBC) with Energy-Based Models (EBMs) to better capture this multimodality. We propose Data-Augmented IBC (DA-IBC), which improves learning by perturbing expert actions to form the counterexamples of IBC training and using better initialization for derivative-free inference. 
% In experiments on the CARLA simulator with Bird's-Eye View inputs and no route input, DA-IBC outperforms BC and IBC in urban driving tasks. 
Experiments in the CARLA simulator with Bird's-Eye View inputs demonstrate that DA-IBC outperforms standard IBC in urban driving tasks designed to evaluate multimodal behavior learning in a test environment. 
The learned energy landscapes are able to represent multimodal action distributions, which BC fails to achieve.
% Experiments in the CARLA simulator with Bird's-Eye View inputs demonstrate that DA-IBC outperforms both BC and standard IBC in routeless urban driving tasks designed to evaluate multimodal behavior learning in a test environment, with
% the learned energy landscapes able to represent multimodal action distributions.
% and trajectories 
% confirm the agent’s ability to represent multimodal behaviors in ambiguous scenarios.
\end{abstract}
     
% \begin{resumo} 
%   Este meta-artigo descreve o estilo a ser usado na confecção de artigos e
%   resumos de artigos para publicação nos anais das conferências organizadas
%   pela SBC. É solicitada a escrita de resumo e abstract apenas para os artigos
%   escritos em português. Artigos em inglês deverão apresentar apenas abstract.
%   Nos dois casos, o autor deve tomar cuidado para que o resumo (e o abstract)
%   não ultrapassem 10 linhas cada, sendo que ambos devem estar na primeira
%   página do artigo.
% \end{resumo}

% Multi-Modal Trajectory Prediction with Energy-Based Models
% Deep Imitative Models for Flexible Inference, Planning, and Control
% Controlling Steering with Energy-Based Models
% Model-Based Planning with Energy-Based Models

% 
\section{Introduction}
Many learning-based approaches for autonomous driving rely on Behavior Cloning (BC) \cite{Pomerleau1991}, a supervised method that learns from offline expert demonstrations \cite{nvidiabc,videobc,chauffernet,codevilla2018endtoend,codevilla2019exploring}. In BC, a human driver provides input observations paired with corresponding control commands.
For example, Codevilla \cite{codevilla2018endtoend,codevilla2019exploring} applied BC for autonomous driving in the CARLA simulator. A large dataset of human driving data was collected and augmented using image processing techniques to train end-to-end policies conditioned on the desired route. 
%Further improvements were introduced in \cite{ral_prob_bc}, where a deep ResNet network extracted features from multimodal sensor data (cameras, LiDAR, and radar) to build comprehensive feature maps, while a probabilistic motion planner addressed trajectory uncertainties.

In autonomous driving, multimodality describes the presence of multiple valid control choices or trajectories available to a driver under the same circumstances. For example, at an intersection, a vehicle might continue forward, make a left turn, or turn right, depending on elements like surrounding traffic and pedestrian activity. %\cite{codevilla2018endtoend}. 
Similarly, when overtaking, a driver can opt for different acceleration profiles or merge timings based on the current traffic flow \cite{chauffernet}. Braking behavior also varies—some drivers may begin slowing down smoothly and early, while others may delay braking and decelerate more abruptly.
% \citep{chai2019multipath}.
%
% 
Traditional BC can not capture such variability, as it typically learns a deterministic policy that maps each observation $\mathbf{o}$ to a single action $\hat{\mathbf{a}}$, as in
$\hat{\mathbf{a}} = F_\theta(\mathbf{o})$.
When the demonstration data includes multiple valid behaviors for the same input, BC averages these actions, resulting in mode collapse, a failure mode where the learned policy does not reflect the diversity of real-world driving decisions. %\citep{rhinehart2018deep}. 
For instance, if demonstrations include both left and right turns in similar contexts, the model may output a steering command between the two, causing the vehicle to drive straight and potentially fail. 
% Figure~\ref{fig:unimodality} illustrates this challenge in a multimodal scenario where the desired trajectory is not specified. 
In this context, a unimodal policy, such as the one learned by BC, captures only one possible behavior, while a multimodal policy can represent multiple valid actions for the same sensory input.

% Although Behavior Cloning (BC) \cite{Pomerleau1991} remains one of the most straightforward supervised learning approaches for acquiring robotic skills in real-world environments, often with remarkable results \cite{Zhang2018a}, it is not a suitable way to learn multimodal action distributions that can represent multiple suitable strategies at the same driving situation. 

Energy-Based Models (EBMs) offer a compelling alternative for capturing the inherently multimodal nature of driving behavior. Rather than directly predicting an action, EBMs learn an energy landscape over the action space, where desirable actions are associated with lower energy values and less favorable ones with higher energy.
This formulation naturally accommodates multiple plausible actions by assigning low energy to all feasible choices, avoiding the need to commit to a single output. As a result, EBMs can represent diverse, human-like driving responses without collapsing to an average of conflicting behaviors.
In contrast to BC, which minimizes a supervised loss on expert actions, EBMs are trained using a contrastive approach, encouraging low energy for expert demonstrations and high energy for non-expert samples. This learning strategy enhances generalization and helps the model reject unrealistic actions during inference.

% BC frames the imitation of expert demonstrations as a supervised learning problem, and despite well-founded concerns 
% %(both empirical and theoretical) 
% regarding its limitations (e.g., compounding errors \cite{Ross2010}), in practice, it enables some of the most remarkable results in real-world robotics, allowing robots to generalize complex behaviors to novel and unstructured scenarios \cite{Zhang2018a}.  
%
% Although substantial research efforts have focused on developing new imitation learning techniques \cite{Ho2016,Abbeel2004} to overcome BC’s known weaknesses, 
In this work, we follow the implicit behavioral cloning approach \cite{Florence2022}, 
% which changes the structure of the policy itself.
% Instead of using BC policies that are commonly modeled using explicit, continuous feedforward architectures (deep networks) in the form  $\hat{\mathbf{a}} = F_\theta(\mathbf{o})$,  
% which directly associate input observations $\mathbf{o}$ with output actions $\mathbf{a} \in A$, 
adopting a reformulation of BC using implicit models:
%\cite{Florence2022}:
% More specifically, the composition of $\arg\min$ with a continuous energy function $E_\theta$ (see Section 2 for details) to define the policy $\pi_\theta$:
$
\hat{\mathbf{y}} = \arg\min_{\mathbf{y} \in Y} E_\theta(\mathbf{x},\mathbf{y})
$
instead of
$ \hat{\mathbf{y}} = F_\theta(\mathbf{x}). $
This formulation casts imitation as a conditional energy-based modeling problem 
%\cite{LeCun2006}  
%(Fig. 1)  
and, at inference time (given $\mathbf{x}$), performs implicit regression by optimizing the best action $\hat{\mathbf{y}}$  
through sampling or gradient-based optimization \cite{Du2019}.  
%
% As partial components (e.g., value functions) for various reinforcement learning (RL) methods [13, 14, 15, 16], our work presents a distinct yet simple approach: performing BC with implicit models. Furthermore, this enables a unique case study investigating the choice between implicit versus explicit policies, which could inform other policy learning contexts beyond BC.
%
Findings from \cite{Florence2022} demonstrate that implicit models for BC can learn long-horizon, closed-loop visuomotor tasks more effectively than their explicit counterparts.
% —and, surprisingly, introduce a novel class of BC baselines that rival state-of-the-art \textit{offline} reinforcement learning algorithms in standard simulated benchmarks. 
Their success is partially attributed to their capacity to represent not only multimodal distributions but also discontinuous functions.  
While implicit models for BC have been applied to robotic behavior learning \cite{Florence2022}, in this work, we propose their use in scenarios involving multimodal action selection for autonomous vehicles navigating urban environments, specifically leveraging the high-fidelity autonomous driving simulator CARLA \cite{Dosovitskiy2017}.
% Prakash et al. (2021) applied EBMs to trajectory prediction, capturing multiple plausible futures states in traffic scenes. 
% Rhinehart et al. (2019) introduced Deep Imitative Models to generate expert-like trajectories for planning and control. 
% Unlike these works, which focus on future prediction or general imitation, our approach applies IBC to routeless urban driving in CARLA. 
%We propose Data-Augmented IBC (DA-IBC), which improves counterexample generation and inference initialization, achieving richer multimodal behavior than standard IBC or BC.

% Energy-Based Models (EBMs) have shown strong potential for capturing multimodal behavior in autonomous systems. Prakash et al. (2021) applied EBMs to predict diverse future trajectories in driving scenes, modeling multiple plausible agent behaviors. 
% Rhinehart et al. (2019) introduced Deep Imitative Models, learning expert-like trajectory distributions for flexible planning and control. 

% While these works target future prediction or closed-loop visuomotor tasks, our work applies IBC to routeless urban driving, where action ambiguity is high. We propose Data-Augmented IBC (DA-IBC), which improves training via perturbed expert counterexamples and enhances inference through density-aware initialization. This leads to more expressive multimodal policies than standard IBC or BC in simulated driving environments

Energy-based approaches have been explored to capture multimodality in sequential decision-making. 
% Trajectory Prediction with Latent Belief Energy-Based Model [Prakash et al., 2021] 
In \cite{Pang2021},
pedestrian motion is modeled by combining latent variables with energy functions to generate diverse future trajectories. Although not targeting vehicle control, it demonstrates how EBMs can handle multimodal forecasting in interactive environments. 
Deep Imitative Models \cite{Rhinehart2018}
learn a distribution over expert trajectories and perform inference by optimizing full paths, but rely on maximum likelihood training and trajectory-level reasoning and not on EBMs. 
% Controlling Steering with Energy-Based Models 
The work in \cite{Balesni2023} applies EBMs to control steering in a road following task, but lacks generalization to complex urban settings. In contrast, our proposed IBC method applies EBMs directly to action-level inference in urban driving. 
% , learning to model multimodal action distributions.
% improving multimodal behavior learning through expert-based counterexample sampling and more effective inference initialization

However, conventional IBC has some known limitations when applied to many problems \cite{Singh2023}.
To improve IBC, we propose a new method for generating counterexamples, namely Data-Augmented IBC, which draws action samples from the expert set and adds a perturbation to each one to form the set of counterexamples, making learning in IBC more effective. 
This makes the negative sampler better since it generates more realistic counterfactual actions than randomly sampling over the complete action space as done in standard IBC.
We evaluate this new method and show its effective multimodal behavior learning in an autonomous driving setting where the route to follow is unknown to the agent.
Our main contributions are: 1) we improve IBC training through expert-based counterexample sampling; 2) more effective inference initialization and sampling; 3) we apply the proposed method in the realistic CARLA simulator for urban driving, showing that the energy function learns to model different actions in the same situations.

\section{Methods}

%%%%%%%%%%%%%%%%%%%%
\subsection{Bird's-Eye View - BEV representation}
\begin{figure}[thpb]
  \centering
  \subfigure[Frontal Cameras]{
    \label{fig:cameras}  
    {\includegraphics[scale=0.42]{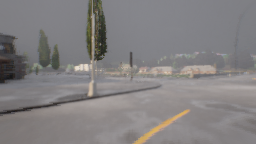}}
    {\includegraphics[scale=0.42]{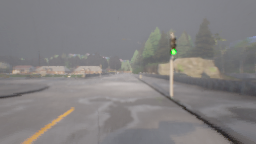}}
    {\includegraphics[scale=0.42]{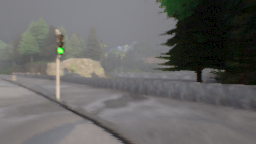}}
  }
  % \subfigure[Sparse trajectory]{
  % {\includegraphics[scale=0.3]{images/traj_plot.png}}
  % \label{fig:traj_plot} 
  % }
  % \\
   \subfigure[five-channel BEV image]{ 
    {\includegraphics[scale=0.35]{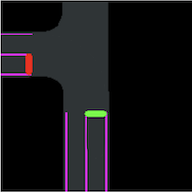}}
    % - Eric. TODO: I edited this image for now, removing the route (need to render it properly though)
    \label{fig:bev_final}
  }
  \subfigure[
  % Bird's-Eye View (BEV) channels: 
  % route, 
  drivable area,
  and lane boundaries
  ]{
    % {\includegraphics[scale=0.35]{images/desired_route.png}}
    % \quad
    {\includegraphics[scale=0.35]{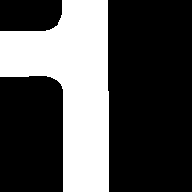}}
    \quad
    {\includegraphics[scale=0.35]{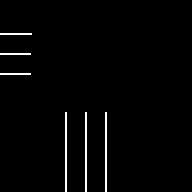}}
    % \quad     
    % \qquad \qquad
    % \qquad
    \label{fig:bev}
  }
  \subfigure[BEV's traffic lights channels at timesteps (-1, -9, -16)]{
    {\includegraphics[scale=0.35]{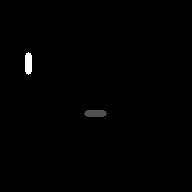}}
    \quad
    {\includegraphics[scale=0.35]{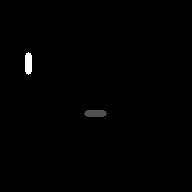}}
    \quad
    {\includegraphics[scale=0.35]{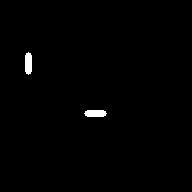}}
    \label{fig:bev_lights}
    %\quad
    }
  \caption{
  a) Images from three frontal cameras positioned on the left, center, and right side of the vehicle. These images were taken after the initial interactions of the agent in the CARLA simulation environment. 
  % Each camera captures a 256$\times$144 RGB image.
  The corresponding 
  bird's-eye view channels used as inputs to our agent are shown next:
  % in (b), (c) and (d), computed at the same moment as shown in (a).
%
% b) The corresponding sparse trajectory visual input is captured in the same frame. The points from the sparse trajectory and the vehicle's highlighted position are depicted as circles with a 10-pixel radius, using the same scale (pixels per meter) and perspective as the bird's-eye view (BEV) representation. 
%When fed into the CGAN, this image is represented with a single channel and has a size of 192x192 pixels.
%
b)
The final BEV image displays all five channels combined in different colors;
c) Two BEV 192$\times$192 channels that represent, from left to right, 
% the desired route, 
the drivable area, and the lane boundaries;
d) BEV's traffic lights channels at timesteps (-1, -9, -16), i.e., representing the past states of the traffic light channel (grey color is red light off; white color means red ligh on).
  }
    %These images are fed to the networks as they are.  
  \label{fig:cameras_all}
\end{figure}

%%% from our other paper: (change a bit the text below)
%
The bird's-eye view of a vehicle represents its position and movement in a top-down coordinate system 
\cite{chauffernet}. The vehicle's location, heading, and speed are represented by $p_t$, $\theta_t$, and $s_t$ respectively. 
The top-down view is defined so that the agent's starting position is always at a fixed point within an image (the center of it). 
% The environment is represented by a set of images with a certain resolution,
%
Furthermore, it is represented by a multi-channel image, where each channel is a binary map indicating the presence or absence of a specific semantic class (e.g., road, vehicle, pedestrian), with values restricted to 0 or 1, and rendered at a ground sampling resolution of $\phi$ meters per pixel.
The BEV of the environment moves as the vehicle moves, allowing the agent to see a fixed range of meters in front of it. 
% For instance, the BEV representation for the vehicle whose three frontal cameras are shown in Fig.~\ref{fig:cameras} is given in Fig.~\ref{fig:bev}, where the desired route, drivable area and lane boundaries are shown as three BEV channels. 
For instance, the BEV representation for a vehicle located just before a traffic light is given in Fig.~\ref{fig:bev}, where 
% the desired route, 
drivable area and lane boundaries are shown as 
% three
two BEV channels.
Besides, other three BEV channels for the traffic lights representing the state of the traffic lights (green or red) at different timesteps are shown in Fig.~\ref{fig:bev_lights}. 
Fig.~\ref{fig:bev_final}
shows the integrated five-channel BEV image, with each channel rendered with a different color. 
% We use a BEV generated by an algorithm that access privileged information from the simulator, but mapping camera images to BEV can be learned \cite{Couto2023}.
% Besides, our agents use input BEV images without the route channel.
The mid-level BEV input can be learned as in \cite{Couto2023}, but in this work we use the ground-truth BEV generated using the CARLA simulator.

%%%%%%%%%%%%%%%%%%%%%%%%%%%%%%%%

%%%%%%%%%%%%%%%%%%%%

\subsection{Behavior Cloning (BC)}
Behavior cloning is a supervised imitation learning method that trains a policy to mimic expert behavior. In our work, a CNN learns to output two continuous actions—acceleration and steering—from a Bird's-Eye View (BEV) representation and additional state variables.

Let \( \pi_{\theta}(s) \) denote the policy mapping state \( x \) to action \( y \), and \(\mathcal{D} = \{(x_i, y_i)\}_{i=1}^N\) be the expert dataset. Rather than using a standard MSE or cross-entropy loss, we model a stochastic policy with a Beta distribution, which naturally handles variables bounded in \([-1, 1]\).
For each sample \(i\), with actions \(y_i = (y_{i}^\mathrm{acc}, y_{i}^\mathrm{steer})\) and policy outputs \(\pi_{\theta}(x_i) = (\alpha_{i}^\mathrm{acc}, \beta_{i}^\mathrm{acc}, \alpha_{i}^\mathrm{steer}, \beta_{i}^\mathrm{steer})\), we define the Beta density as:
\[
f(z; \alpha, \beta) = \frac{z^{\alpha-1} (1-z)^{\beta-1}}{B(\alpha, \beta)},
\]
with \(B(\alpha, \beta)\) as the normalization constant. Actions are rescaled from \([-1, 1]\) to \([0, 1]\) via \( y_i' = \frac{y_i + 1}{2} \).
The negative log-likelihood loss is then:
% \[
% \mathcal{L}(\theta) = -\frac{1}{N} \sum_{i=1}^N \left[ \log f\left( \frac{a_{i}^\mathrm{acc} + 1}{2}; \alpha_{i}^\mathrm{acc}, \beta_{i}^\mathrm{acc} \right) + \log f\left( \frac{a_{i}^\mathrm{steer} + 1}{2}; \alpha_{i}^\mathrm{steer}, \beta_{i}^\mathrm{steer} \right) \right].
% \]
%
\begin{equation}
\mathcal{L}(\theta) = -\frac{1}{N} \sum_{i=1}^N \Big[ \log f\left( \frac{y_{i}^\mathrm{acc} + 1}{2}; \alpha_{i}^\mathrm{acc}, \beta_{i}^\mathrm{acc} \right) 
+ \log f\left( \frac{y_{i}^\mathrm{steer} + 1}{2}; \alpha_{i}^\mathrm{steer}, \beta_{i}^\mathrm{steer} \right) \Big].
\label{eq:bcloss_car}
\end{equation}
% \begin{equation}
% \begin{split}
% \mathcal{L}(\theta) = -\frac{1}{N} \sum_{i=1}^N \Big[ &\log f\left( \frac{y_{i}^\mathrm{acc} + 1}{2}; \alpha_{i}^\mathrm{acc}, \beta_{i}^\mathrm{acc} \right) \\
% &+ \log f\left( \frac{y_{i}^\mathrm{steer} + 1}{2}; \alpha_{i}^\mathrm{steer}, \beta_{i}^\mathrm{steer} \right) \Big].
% \end{split}
% \label{eq:bcloss_car}
% \end{equation}
%
This approach lets the model capture uncertainty by learning the distribution parameters instead of fixed action values.

\subsection{Implicit Behavior Cloning (IBC)}
As defined in \cite{Florence2022}, we consider an implicit model to be any composition  
$\left(\arg\min_y \circ E_\theta(x,y)\right)$  
in which inference is performed using a general-purpose function approximator  
$(E: \mathbb{R}^{m+n} \rightarrow \mathbb{R}^1)$  
to solve the optimization problem  
$[ \hat{y} = \arg \min_y E_\theta(x,y)]$.  
We leverage techniques from the energy-based model (EBM) literature to train such a model. Given a dataset of samples $(\{x_i, y_i\})$ and regression bounds  
$(y_{\mathrm{min}}, y_{\mathrm{max}} \in \mathbb{R}^m)$,  
training consists of generating a set of negative counterexamples  
\textcolor{red}{
$\{\tilde{y}_{i}^j\}_{j=1}^{N_{\mathrm{neg}}}$  
}  
for each sample $(x_i)$ in a batch and employing an InfoNCE-style loss function \cite{Oord2018}. This loss corresponds to the negative log-likelihood of  
$[ p_\theta(y|x) = \frac{\exp(-E_\theta(x,y))}{Z(x,\theta)}]$,  
where the counterexamples are used to estimate  
$Z(x_i, \theta)$:  
\vspace{-0.5cm}
\begin{equation}
\mathcal{L}_{\mathrm{InfoNCE}}(\theta) = \sum_{i=1}^N -\log \left( \tilde{p}_\theta(y_i \mid x_i, \textcolor{red}{\{\tilde{y}_{i}^j\}_{j=1}^{N_{\mathrm{neg}}}} ) \right),
\label{eq:ibcloss}
\end{equation}
\begin{equation}
% \[
\tilde{p}_\theta(y_i \mid x_i,  
\textcolor{red}{\{\tilde{y}_{i}^j\}_{j=1}^{N_{\mathrm{neg}}}} ) = \frac{e^{-E_\theta(x_i, y_i)}}{e^{-E_\theta(x_i, y_i)} +  
	\textcolor{red}{\sum_{j=1}^{N_{\mathrm{neg}}}} e^{-E_\theta(x_i,  
		\textcolor{red}{\tilde{y}_{i}^j}) }}.
% \]
\label{eq:prob_ebm}
\end{equation}
Thus, the lower the energy $E_\theta(x, y)$, the higher the probability of the action $y$, conditioned on the observation $x$.
With a trained energy model  
$ E_\theta(x,y) $,  
implicit inference can be performed via stochastic optimization to solve  
\(\hat{y} = \arg \min_y E_\theta(x,y)\) (see Section ~\ref{sec:dfo}).
% \cite{Florence2022}.  

Moreover, this implicit model allows for the representation of probability distributions over actions with multiple modes, meaning that more than one action may be appropriate for a given context (for the same input image frame).  
    In contrast, conventional (explicit) BC does not support this, as it models only unimodal distributions (such as a Gaussian distribution). If such a model were placed in a scenario requiring multimodal action selection, the policy would likely choose an action that represents the mean of all modes. This could lead a vehicle to drive straight ahead when it should turn left or right, resulting in an invalid or unsafe action.

\subsection{Data-Augmented IBC (DA-IBC)}
In \cite{Florence2022}, for each observation \( x_i \) in a batch, the set of negative counter-examples \( \{ \tilde{y}_j^i \}_{j=1}^{N_{\text{neg}}} \) is generated by uniformly sampling from the action space: \( \tilde{y}_j^i \sim U(y_{\min}, y_{\max}) \). 
In this work, instead of uniform sampling, we generate negative counter-examples by sampling with replacement from the expert set\footnote{The sampling is actually made from a random minibatch of actions from the expert dataset.} and subsequently adding Gaussian noise drawn from \( \mathcal{N}(0, \sigma^2 I) \) to introduce perturbations. We refer to our method as \textit{Data-Augmented IBC (DA-IBC)}.
The set \( \{ \tilde{y}_j^i \}_{j=1}^{N_{\text{neg}}} \) is generated randomly and dynamically within the training loop for each observation \( x_i \) in a batch \( \mathcal{B} = \{(x_i, y_i)\} \) from the expert dataset \( \mathcal{D} \). As a result, different sets of counter-examples are created in each training iteration. 
Note that the actions in the set \( \{ \tilde{y}_j^i \}_{j=1}^{N_{\text{neg}}} \) are not dependent on any particular value of $x_i$, unlike \cite{Singh2023} for instance.
%More specifically, the set of all action values in the expert set is sampled 
%all expert actions corresponding to a given observation \( x_i \) are collected into a set. 
If the number of available expert actions is smaller than \( N_{\text{neg}} \), we sample with replacement until the desired number is reached. 
Finally, each action in the set is perturbed with Gaussian noise to form the negative counter-examples.

\subsubsection{Derivative-Free Optimization (DFO)}
\label{sec:dfo}
We use an adapted version of DFO from \cite{Florence2022}, shown in Algorithm \autoref{algo:dfo}, to perform implicit inference \(\hat{y} = \arg \min_y E_\theta(x,y)\). 
% with a trained EBM network.
%
In DFO, the samples $\{\tilde{y}_i\}$  are initialized by sampling from a categorical distribution of the expert actions with probabilities equal to the corresponding weights $\{\tilde{w}_i\}$, with
\( w_i = \frac{1}{\hat{g}(y_i^{\text{expert}})} \)
obtained by KDE (Section~\ref{sec:KDE}).
After the $N_{\text{iters}}$ iterations of the method,
instead of returning the action with maximum probability, a sample is returned by drawing from a categorical distribution of the actions 
$\{\tilde{y}_i\}$
and its respective probabilities 
$\{\tilde{p}_i\}$
found in the last iteration. 
This modification enhances the agent's potential to perform multimodal behavior.
% This change will make possible for the agent to have a higher chance of exhibiting multimodal behavior.

% iter_std=0.22,  \sigma_{\text{init}}
% K = 0.5

\begin{algorithm}[tb!]
\caption{Derivative-Free Optimizer}
\begin{algorithmic}[1]
% {\small
    \State \textbf{Result:} $\hat{y}$
    \State \textbf{Initialize:}   
    \textcolor{red}{
    $\{\tilde{y}_i\}_{i=1}^{N_{\text{samples}}} \sim \text{Categorical}(\{y_i^{\text{expert}}\}, \{w_i\})$, \quad
    $w_i \propto \text{KDE}(y_i^{\text{expert}})$, \quad $\sigma = \sigma_{\text{init}}$
    }
    \For{$\text{iter} = 1, 2, \dots, N_{\text{iters}}$}
        \State
        $\{E_i\}_{i=1}^{N_{\text{samples}}} = \{E_{\theta}(\mathbf{x}, \tilde{y}_i)\}_{i=1}^{N_{\text{samples}}}$ {\scriptsize (Compute energies);}
        \State 
        $\{\tilde{p}_i\}_{i=1}^{N_{\text{samples}}} = \left\{\frac{e^{-E_i}}{\sum_{j=1}^{N_{\text{samples}}} e^{-E_j}}\right\}$ {\scriptsize (Softmax probab.);}

        \If{$\text{iter} < N_{\text{iters}}$}
            \State 
            $\{\tilde{y}_i\}_{i=1}^{N_{\text{samples}}} \sim \text{Multinomial}(N_{\text{samples}}, \{\tilde{p}_i\}_{i=1}^{N_{\text{samples}}},$
            \State \qquad \qquad \hspace{0.1cm} 
            $\{\tilde{y}_i\}_{i=1}^{N_{\text{samples}}})$ {\scriptsize (Resample with replacement);}
            
            \State 
            $\{\tilde{y}_i\}_{i=1}^{N_{\text{samples}}} \gets \{\tilde{y}_i\}_{i=1}^{N_{\text{samples}}} + \mathcal{N}(0, \sigma)$ {\scriptsize(Add noise);}
            \State
            $\{\tilde{y}_i\}_{i=1}^{N_{\text{samples}}} = \text{clip}(\{\tilde{y}_i\}_{i=1}^{N_{\text{samples}}}, y_{\min}, y_{\max})$
            \State 
            $\sigma \gets K \sigma$ {\scriptsize (Shrink sampling scale);}
        \EndIf
    \EndFor

    % \State  
    % \textcolor{red}{
    % $\hat{y} \sim \text{Categorical}(\{\tilde{y}_i\}, \{\tilde{p}_i\})$
    % }
    \State \textbf{Return:} 
    \textcolor{red}{$\hat{y} \sim \text{Categorical}(\{\tilde{y}_i\}, \{\tilde{p}_i\})$}
% } % <--- add this    
\end{algorithmic}
\label{algo:dfo}
\end{algorithm}

\subsection{Kernel Density Estimation (KDE)}
\label{sec:KDE}

% A Kernel Density Estimator (KDE) is a non-parametric way to estimate the probability density function (PDF) of a random variable.
% %
% For each data point in the sample, the kernel function is centered at that point. The KDE is then computed by summing up all these kernel functions and normalizing by the number of data points. 
Mathematically, the KDE \cite{silverman1986density}
for a point \(x\) is given by:

\begin{equation}
   \hat{g}(x) = \frac{1}{N h} \sum_{j=1}^{N} K \left( \frac{x - x_j}{h} \right),
    \label{eq:kde}
\end{equation}
where \( \hat{g}(x) \) is the estimated density at point \( x \), \( N \) is the number of data points, \( h \) is the bandwidth, \( x_j \) are the data points, and \( K \) is the kernel function. In particular, we employ the Gaussian kernel
$ K(u) = \frac{1}{\sqrt{2\pi}} e^{-\frac{u^2}{2}}$.

In this work, the KDE is used to weigh the training samples in the BC and IBC losses in (\ref{eq:bcloss_car}) and (\ref{eq:ibcloss}), respectively, and also to form the initial samples in the DFO inference Algorithm  \autoref{algo:dfo} .
By using the inverse of the kernel density estimation of the actions as weights, we give more importance to less frequent actions. 
This approach enhances the model's ability to learn from sparsely represented regions of the action space.
% This approach can help the model learn better on underrepresented parts of the action space. 
%
% The resulting weighted BC loss function can be written as:
% \begin{equation}
% \begin{split}
%  \mathcal{L}(\theta) = -\frac{1}{N} \sum_{i=1}^N w_i \Big[ & \log f\left( \frac{a_{i}^\mathrm{ac} + 1}{2}; \alpha_{i}^\mathrm{ac}, \beta_{i}^\mathrm{ac} \right) \\
%  &+ \log f\left( \frac{a_{i}^\mathrm{steer} + 1}{2}; \alpha_{i}^\mathrm{steer}, \beta_{i}^\mathrm{steer} \right) \Big],
% \end{split}
% \label{eq:weighted_loss}
% \end{equation}
%
% 
The weighted IBC loss is as follows:
\begin{equation}
\mathcal{L}_{\mathrm{InfoNCE}}(\theta) = \sum_{i=1}^N - w_i \log \left( \tilde{p}_\theta(y_i \mid x_i, \textcolor{red}{\{\tilde{y}_{i}^j\}_{j=1}^{N_{\mathrm{neg}}}} ) \right).
\end{equation}
where:
\( w_i = \frac{1}{\hat{g}(y_i)} \);
and \( \hat{g}(y_i) \) is the kernel density estimate evaluated at action \( y_i \).
% The KDE is estimated for the actions in the training set.
% Then the weights $w_i$ are calculated as the inverse of the KDE for each action.
For BC, the loss function is weighted analogously \cite{Antonelo2024}.
% , which is also presented in more detail in \cite{Antonelo2024}.
%
%
In practice, when training the model using stochastic gradient descent or ADAM, instead of directly applying the weighted loss, we employ a sampling strategy that increases the frequency of less common actions in the dataset. Specifically, at each training iteration, a minibatch is drawn using a weighted sampling approach,
%such as PyTorch’s WeightedRandomSampler, 
where the sampling probabilities are defined by the weights, mitigating data imbalance during training. 
% This ensures that less frequent actions are sampled more often, mitigating data imbalance during training.

%This approach ensures that the loss function accounts for the density of actions in the training set, giving higher importance to less frequent actions.

%%%%%%%%%%%%%%%%%%%%%%%%%%%%%%%%%%%%%%%%%%%%%%%%%%

\section{Agent}
\label{sec:agent}

\subsection{Input Representation}
The agent's input 
%\( s \) 
is a 
% six-
five-channel 192×192 bird’s-eye view (BEV) image generated by a CARLA module with city map access. 
This BEV includes 
% three channels for the route, 
two channels for the road and lane, plus three channels encoding traffic light history at time steps -16, -9, and -1.
Additionally, the agent receives the vehicle’s current speed and the previous control actions (acceleration and steering). 
% These scalar inputs are integrated in the first fully connected layer of the network.

\subsection{Action Representation}
% \textbf{TODO: correct, talk about energy}

The vehicle in the CARLA simulator has three actuators: steering \([-1,1]\), throttle \([0,1]\), and brake \([0,1]\). The agent’s action space is \(\mathbf{a} \in [-1,1]^2\), where the two components correspond to steering and acceleration. Negative acceleration implies braking, preventing simultaneous acceleration and braking \cite{Petrazzini2021}.
%
%
% \subsubsection{Network Output}
While the network representing the agent outputs a energy value in IBC, which depends not only on an observation but also on an \textit{action input}, in BC the output models a Beta distribution, which is naturally bounded and avoids manual clipping or squashing \cite{Petrazzini2021}. 
%The Beta distribution also improves handling of extreme maneuvers, such as sharp turns and sudden braking. 
The policy network \(\pi_\theta\) in BC outputs \(\alpha\) and \(\beta\), shaping the Beta distribution to adapt to diverse driving scenarios.

% \subsection{Network Architecture}
% The agent's network (Fig.~\ref{fig:architecture}) maps BEV images and state variables to steering and acceleration actions. 

% A Convolutional Neural Network (CNN) extracts a 256-dimensional embedding from the BEV. Similarly, a Multi-Layer Perceptron (MLP) encodes scalar state variables into a 256-dimensional vector. The concatenated embeddings represent the environment state and are processed by another MLP to predict actions.

% \begin{figure}
%   \includesvg[width=0.45\textwidth]{figures/architecture_scheme.svg}
%   \caption{Network architecture. The BEV input is processed by a CNN, while state variables pass through an MLP. Their embeddings are concatenated and fed into a final MLP that outputs actions. 
%   \textbf{TODO: correct figure } 
%   }
%   \label{fig:architecture}
% \end{figure}

\subsection{Network Architecture}

\begin{figure}[tb]
\centering
% \includesvg[width=0.67\textwidth]{figures/arquitetura_EBM_detailed2.svg}
\includegraphics[width=0.87\textwidth]{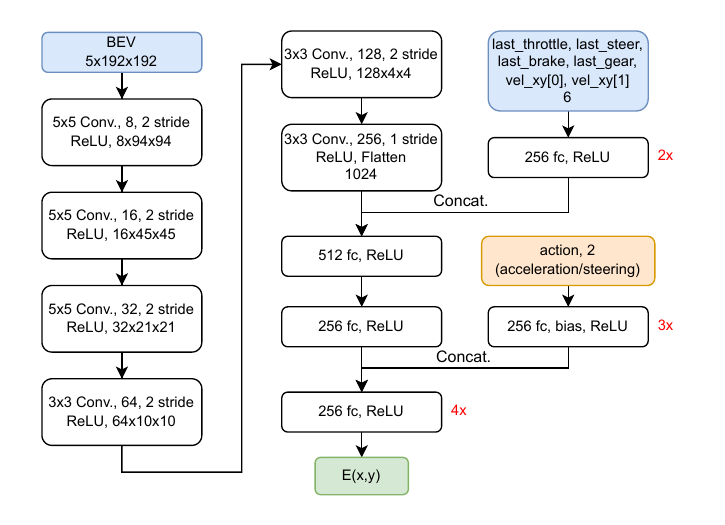}
  \caption{Detailed EBM network architecture. The BEV input has 
  five %six 
  192×192 channels, while the state input consists of previous actions and current speed. The two-dimensional action input corresponds to acceleration and steering values.
  The output $E(x,y)$ is the energy for the observation-action pair $(x,y)$.
  %The policy output consists of \(\alpha\) and \(\beta\) parameters for the Beta distribution, defining both acceleration and steering.
  }
  \label{fig:detail_architecture}
\end{figure}

The goal of the EBM architecture, shown in 
% Fig.~\ref{fig:EBM_architecture}, 
Fig.~\ref{fig:detail_architecture}, 
is to learn the density of actions conditioned on the current BEV, and some state variables. 
The BEV can have a different number of channels, depending on the current experimental setup. 
In our case, it contains all 
five 
% six 
BEV channels from Fig.~\ref{fig:cameras_all}. 

A Convolutional Neural Network (CNN) extracts a 256-dimensional embedding from the BEV. Similarly, a Multi-Layer Perceptron (MLP) encodes scalar state variables into a 256-dimensional vector. 
Together they form the current representation of the environment the agent faces, i.e., the agent's observation. 
The resulting embedding vector for a observation $x$ and an action $y$ are processed by another MLP to map to the energy value $E_\theta(x, y)$. 
% 
% \begin{figure}
% \hspace{-0.25cm}
%   \includegraphics[width=0.52\textwidth]{figures/model_architecture1.png}
%   \caption{EBM Network architecture. The BEV (state variables) is processed by a CNN (MLP) to create an encoding. The image and state embeddings are concatenated, with the resulting vector and the action fed into an final MLP that outputs a single scalar value, the energy.}
%   \label{fig:EBM_architecture}
% \end{figure}
% 
% It is processed by a Convolutionary Neural Network (CNN) that embeds it into a vector of size 256. Similarly the scalar state variables are embedded into a vector of size 256 by a standard MLP. Together they form the current representation of the environment the agent faces. This conditional and an action is processed my another MLP to map to the energy value. 
% 
Fig.~\ref{fig:detail_architecture} shows the detailed network structure.
%%%%%%

%%%%%%%%%%%%%%%%%%%%%%%%%%%%%%%%%%%%%%%%%%%%%%%%%%

\section{Experiments}

% mapa de trajetorias sobrepostas BC x IBC
% mapa da energia mostrando as duas modas - direita esquerda - diferentes instantes umas 4 ou 5
% grafico progresso aprendizado 

% tabelas, metricas BC x IBC (opcional)
In this section, we compare the proposed DA-IBC agent with a traditional BC agent with a Beta policy, presented previously in \cite{Antonelo2024}, and to the conventional IBC agent.

\subsection{Dataset Generation}
%
% - 10 trajectories, first has 740 m, 2000 steps/figure, CARLA Leadboard, town01
% - approx 30 min., 10 traj

\begin{figure}[tb]
  \centering
  \subfigure[\textit{town01}]{
  {\includegraphics[scale=0.4]{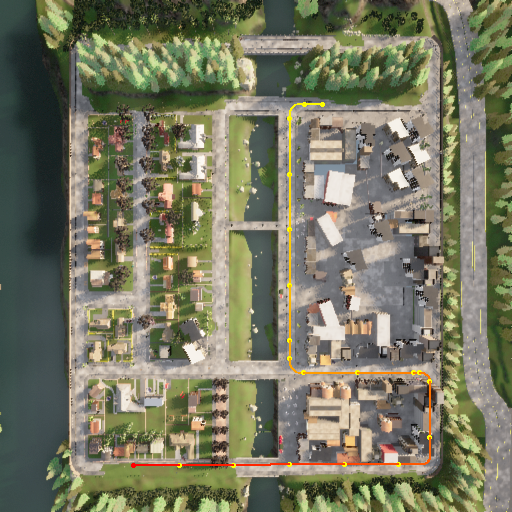}}
  \label{fig:town01}
  }
  % \hspace{0.15cm}
  %
  \subfigure[\textit{town02}]{
  {\includegraphics[scale=0.4]{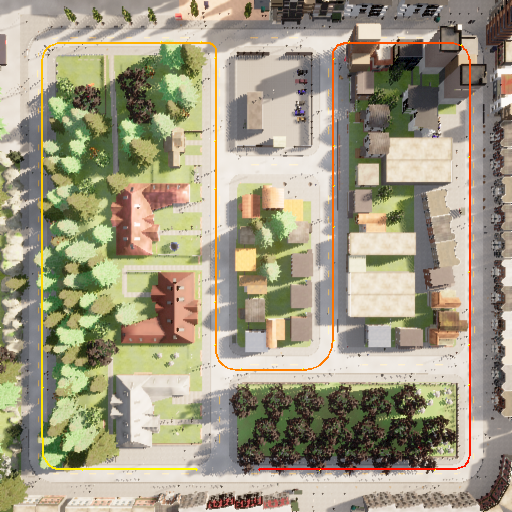}}
  \label{fig:town02}
  }
  \caption{
  (a) The \textit{town01}  environment of the agent includes one of the routes used by the expert to collect data. This highlighted path measures 740 meters
  %, featuring 
  %20 points in the sparse trajectory (indicated by yellow dots) and 
  %762 points in the dense point trajectory (not shown).
  (b) The \textit{town02} is used to evaluate the trained agent in a new environment.
  } 
  \label{fig:towns}
\end{figure}

The expert dataset is sourced from the CARLA Leaderboard evaluation platform \footnote{\url{https://leaderboard.carla.org/}},
% \cite{leaderboard_2020}, 
specifically the \textit{town01} environment with ten predefined trajectories. A deterministic agent, guided by a dense point trajectory and a PID controller \cite{chen2019learning}, generates the dataset. The expert follows a dense trajectory for precise navigation and obeys traffic signals, stopping at red lights and accelerating on green. 
% The learning agent has no route information as input so that multimodal behavior can be evaluated more easily.
%the agent only has access to a sparse trajectory for general guidance. The expert also obeys traffic signals, stopping at red lights and accelerating on green.
%
Fig.~\ref{fig:town01} illustrates one of the ten routes used for training. This trajectory, invisible to the agent, is shown as a gradient line from yellow to red.
%, while the sparse trajectory consists of yellow dots spaced 50 meters apart or placed at movement transitions (e.g., turns).

The dataset was recorded at 10 Hz, yielding 10 observation-action pairs per second. The shortest route contains 1480 samples (2.5 minutes), with an average route having 2129 samples (3.5 minutes). In total, the dataset comprises 21,287 labeled samples (30 GB), covering approximately 8 km or 36 minutes of driving.

%==============================================
%==============================================

\subsection{Settings}

% Tables~\ref{tab:hyperparamsBC}, \ref{tab:hyperparamsEBM}, \ref{tab:hyperparamsDFO}, and \ref{tab:hyperparamsKDE} show the hyperparameter values for the BC, EBM, EBM inference DFO, and KDE algorithms, respectively.

All agents were trained with ADAM step size of 
$10^{-5}$, for a maximum of 400 epochs, using minibatch size of $N = 128$.
For EBM training, $N_{\mathrm{neg}} = 1024$.
For DFO inference,
$N_{\text{samples}} = 16384$,
$\sigma_{\text{init}} = 0.5$,
$N_{\text{iters}}=5$, and
$K=0.5$.
%
% For KDE, the kernel bandwidth was set to $h=0.2$ by trial and error. This is a sensitive parameter.
The kernel bandwidth for KDE was set to $h = 0.2$ through trial-and-error experimentation, given the sensitivity of this parameter and its strong influence on the smoothness of the estimated density function. Additionally, Python and Pytorch were used to implement the networks and their training procedures.

\subsection{Results}
\begin{figure}[tb]
    \centering
    \includegraphics[width=0.8\linewidth]{
    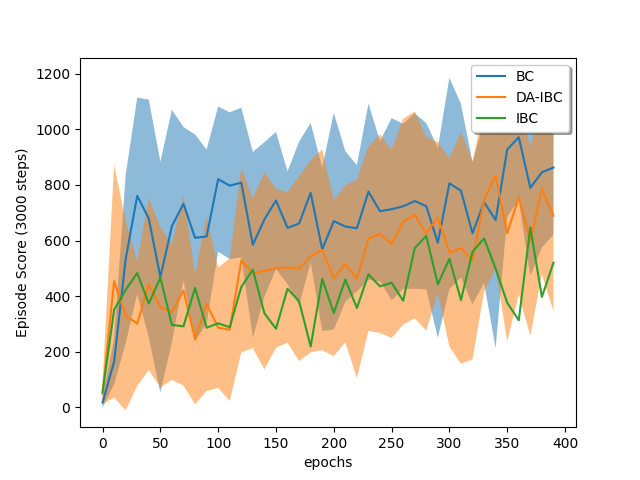
    }
    \caption{Training progress of agents BC, IBC and DA-IBC with training data from \textit{town01} and evaluation in \textit{town02}. 
    The score is the distance traveled multiplied by the \textit{traffic light penalty} ($0.7^{n_i}$, where $n_i$ is the number of infractions in the episode of $300s$), and it
 primarily reflects safe driving behavior and not explicitly multimodality.
    For each method, the average and standard deviation are computed over three different agents 
    evaluated ten times for $3,000$ steps (or $300s$).
    This evaluation is done each 10 training epochs.
    Note that both BC and IBC methods are offline training methods, and the above plot serves to monitor the intermediate steps of learning.
    }
    \label{fig:train-progress}
\end{figure}

%%%%%%%%%%%%%%%%%%%%%%%%%%%%%%%%%%%%%%%%%%%%%%%%
We trained agents with data collected in \textit{town01} based on three methods, the proposed DA-IBC, conventional IBC, and BC with the outputs modeling a Beta distribution. The evaluation was done in \textit{town02} CARLA environment (Fig.~\ref{fig:town02}), with the episode score computed as the traveled distance multiplied by the traffic light penalty. This score only shows the ability to drive without infractions, not taking into account action multimodality.
In Fig.~\ref{fig:train-progress}, this score is shown for all three methods, where the average was computed for three agents per method and over  10 episodes each of $3,000$ timesteps for each agent.
% \footnote{This experiment will be executed at least for more one agent. In case the paper is accepted, the plot in Figure 4 will be updated with these new data. The experiment takes a long time, because every ten epochs require an evaluation in the simulator for each agent.}.
%
% Even though BC attains higher score faster than IBC, its learning problem is easier when compared to IBC, which needs to learn an energy landscape as a function of the action instead of the average action taken at a particular observation as done in BC. Thus, it takes more time to attain higher scores for IBC agents. In addition, IBC learns multimodal action distributions, allowing multimodal behavior when generating vehicle trajectories, which is impossible for BC. 
BC achieves higher scores more quickly, but its task is easier: it predicts the average action for each observation, while IBC must learn an energy landscape over actions.
% Notice that this ability is not taken into account in the performance metrics of the training progress.

%%%%%%%%%%%%%%%%%%%%%%%%%%%%%%%%%%%%%%%%%%%%%%%%
%%%%%%%%%%%%%%%%%%%%%%%%%%%%%%%%%%%%%%%%%%%%%%%%

\begin{figure}[tb]
    \centering
    \includegraphics[width=0.47\linewidth]{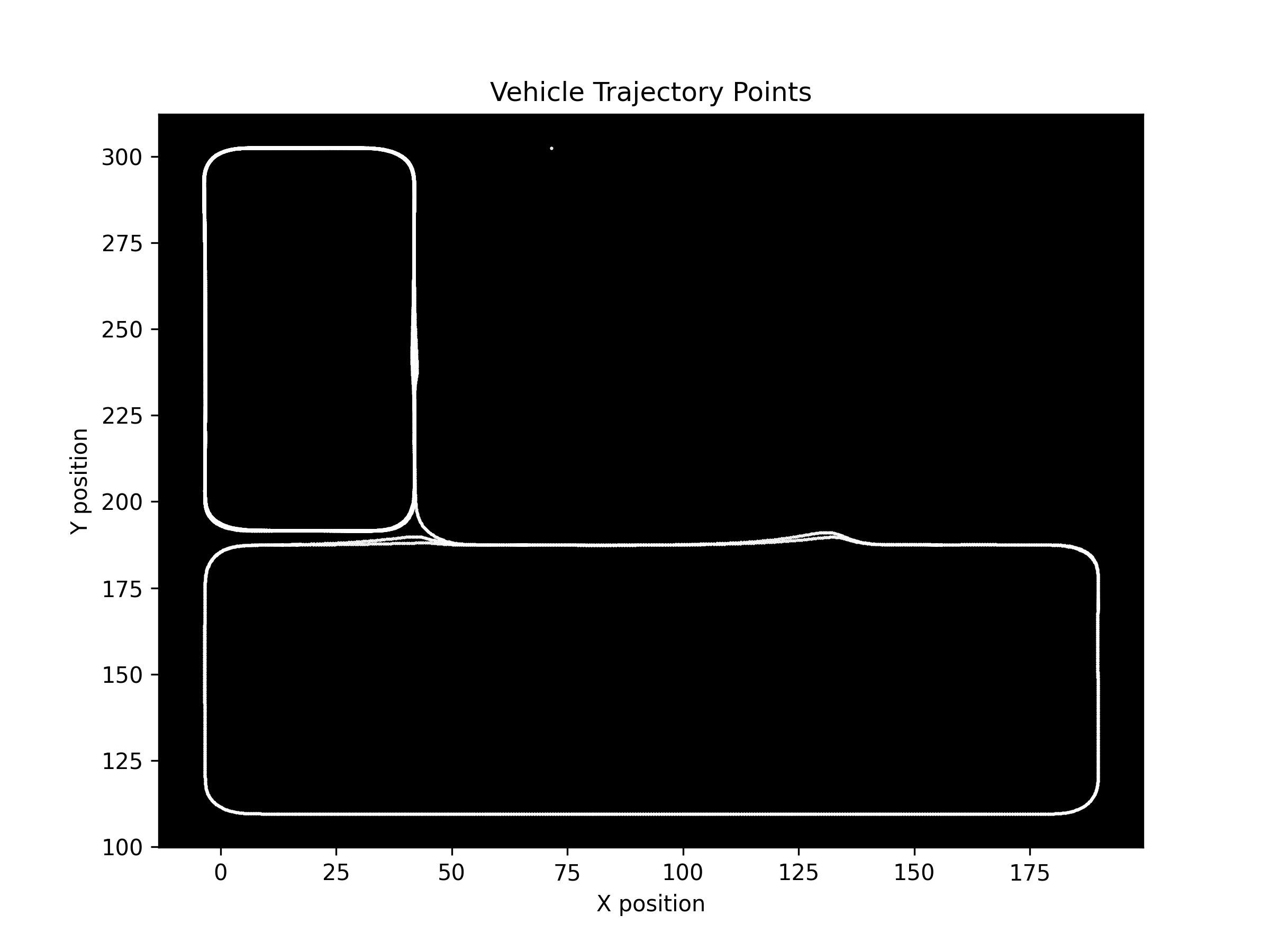} 
    \includegraphics[width=0.47\linewidth]{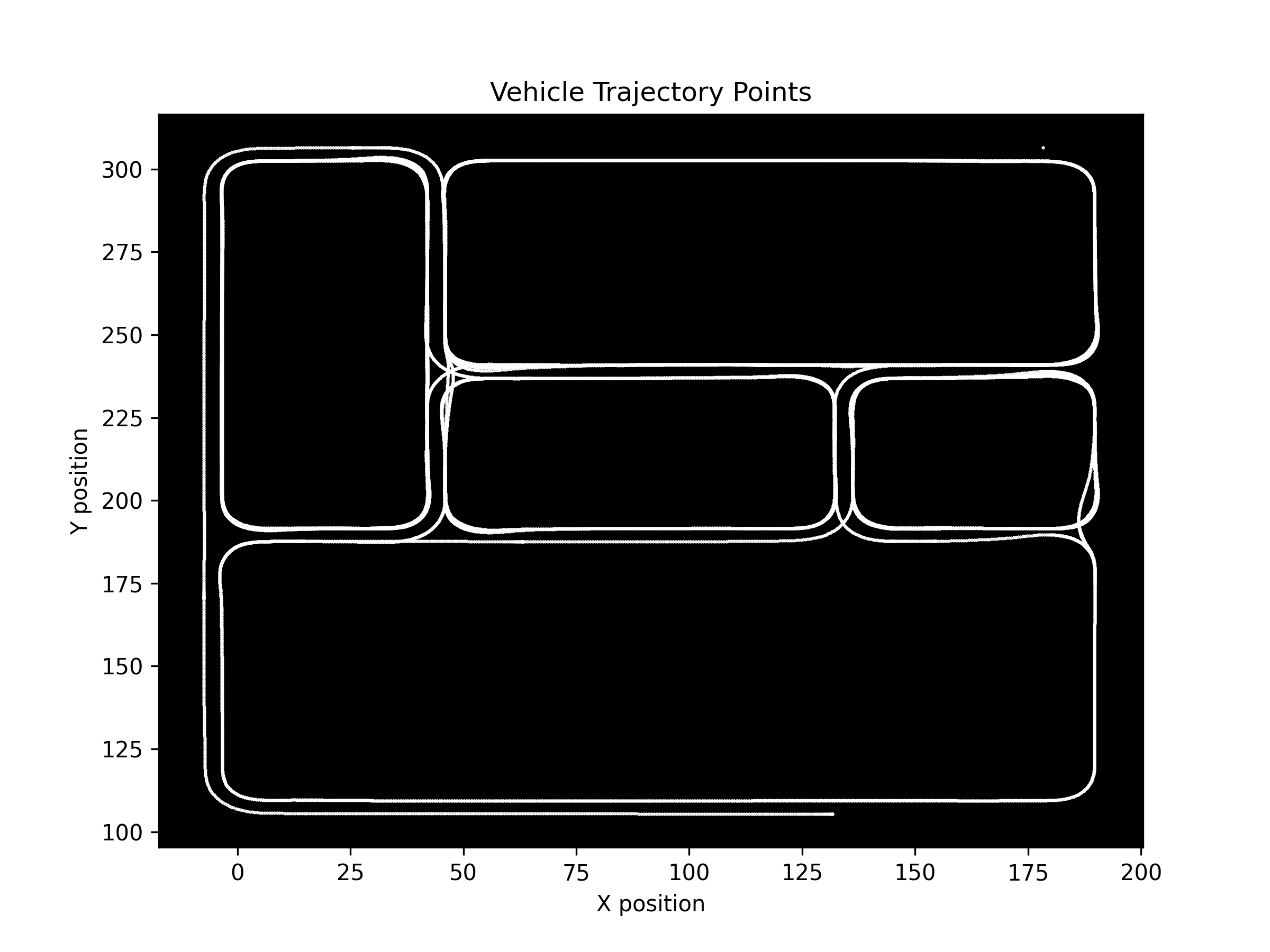} \\
    \small BC \qquad \hspace{5.5cm} DA-IBC
    \caption{
    Trajectory of agents during a simulation of $3,000s$ in \textit{town02}, without a route to follow.
    Left:
    The BC agent can not learn multimodal behavior, being stuck into one or two blocks (likely due to noise) of the city. 
    % It was able to get out of one block likely due to noise and not to multimodality.
    Right:
    The DA-IBC agent learned multimodal action distributions, being able to make different decisions at the same intersections of the city. %and, consequently, visit the whole city map.
    }
    \label{fig:res:bc_long}
    \label{fig:res:ibc_long}
\end{figure}

% We evaluated trained BC and DA-IBC agents in a new \textit{town02} environment for $3,000$ timesteps to test multimodal decision making. 
In order to evaluate multimodal decision making, we let both BC and DA-IBC agents  navigate freely in \textit{town02}, after their training with data from \textit{town01}.
% In Fig.~\ref{fig:res:bc_long}, the trajectory of the BC agent only visits two blocks of the city not due to action multimodality, but to noise from the simulation during decision at a T-intersection, while in Fig.~\ref{fig:res:ibc_long}, the DA-IBC agent can visit all the blocks of the city because of the multimodal action representation learned by the energy-based model that allows it to choose different actions at the same intersection.
In Fig.\ref{fig:res:bc_long}, the BC agent primarily follows a single trajectory through the town, with occasional deviations caused by stochasticity in the simulator. These deviations do not reflect true multimodal behavior, but rather arise from noise during action execution. 
In contrast,
% , as shown in Fig.\ref{fig:res:ibc_long}, 
the DA-IBC agent learns a genuinely multimodal policy that allows it to take distinct actions at the same decision points, enabling it to explore and cover the entire city. 
Neither simulation resulted in crashes, though some traffic light infractions occurred.
% In both simulations, there were no crashes, but some traffic light infractions occurred.

%%%%%%%%%%%%%%%%%%%%%%%%%%%%%%%%%%%%%%%%%%%%%%%%
%%%%%%%%%%%%%%%%%%%%%%%%%%%%%%%%%%%%%%%%%%%%%%%%
%%%%%%%%%%%%%%%%%%%%%%%%%%%%%%%%%%%%%%%%%%%%%%%%
% \begin{figure}[tb]
%     \centering
%     \includegraphics[width=0.6\linewidth]{images/multi_path_ibc.png}
%     \caption{Trajectory of DA-IBC agent during a simulation of $3,000s$ in \textit{town02}, showing its capacity to learn multimodal action distributions. Without a route to follow, the agent was able to make different decisions at the same intersections of the city and, consequently, visit the whole city map.
%     }
%     \label{fig:res:ibc_long}
% \end{figure}
%%%%%%%%%%%%%%%%%%%%%%%%%%%%%%%%%%%%%%%%%%%%%%%%

% TODO:
% - table with results
% - videos, animations
% 

\begin{figure}[tbh!]
    \centering
%     \includegraphics[width=0.56\linewidth]{eplots/energy_plot_left.png}
% %
%     \includegraphics[width=0.56\linewidth]{eplots/energy_plot_right.png}
    \includegraphics[width=0.9\linewidth]{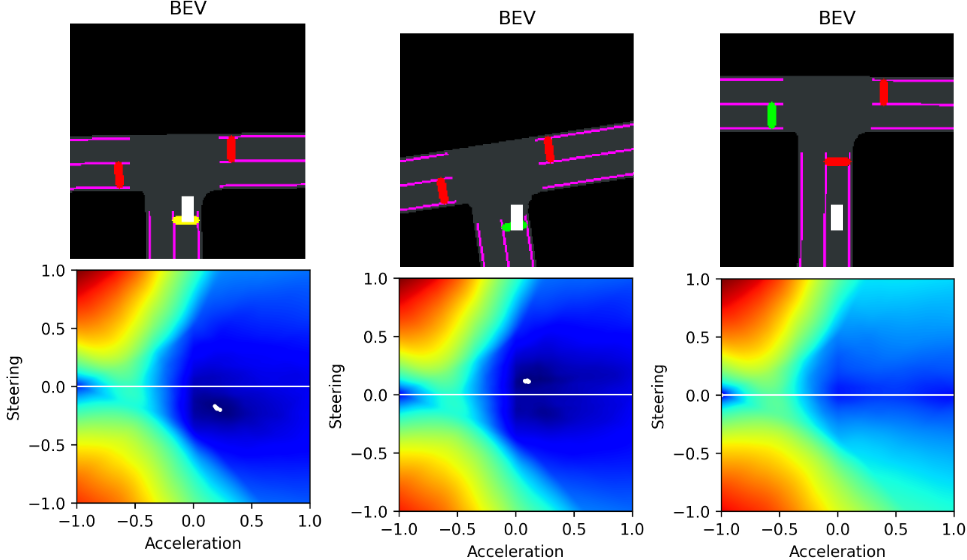}
    \caption{
    Top: the BEV scene as perceived by the agent.
    Bottom: energy \(E(s,a)\) as a function of the two-dimensional action \(a\) (i.e., steering and acceleration).
    A horizontal white line marks the zero steering action.
    The first two columns represent situations where the steering action distribution is multimodal, such as entering an intersection: the minimum in the energy function (blue color) along the steering axis spans a broader range than the third column (stopping at a red light).
    The first column corresponds to a path turning left, 
    while the second one shows the agent turning right at the same intersection.
    Ten DFO inferences were run with the BEV input fixed to the above scene, with corresponding action inferences plotted as small white points in the energy function.
    The latter column shows the steering action distribution nearly unimodal, with 
    %the ten 
    DFO inferences at (0,0).
    % Right:
    % in situations where the steering action distribution is nearly unimodal: 
    % The ten DFO inferences are at (0,0).
    }
    % https://youtu.be/_SLzqnMWtt8  - LEFT
    % https://youtu.be/AhALxsZm7wg - RIGHT
    \label{fig:multimodal-energy}
    \label{fig:multimodal-energy-stop}
\end{figure}

We also investigated the energy landscape of the trained agent in three situations in \textit{town02}. 
In Fig.~\ref{fig:multimodal-energy}, we can see the agent choosing to turn left
\footnote{\url{https://youtu.be/_SLzqnMWtt8}}
or right
\footnote{\url{https://youtu.be/AhALxsZm7wg}}
at the same T-intersection, and the corresponding energy function, where the blue color corresponds to lower energy.
Ten inferences were made with the proposed DFO algorithm and plot as 10 white points in the energy landscape, for each case.
In Fig.~\ref{fig:multimodal-energy-stop}, the energy landscape for the vehicle stopping at the red light indicates a slightly different learned function, with the minimum at (0,0), which signifies zero acceleration and zero steering.
%
% These energy landscapes can work as probabilities distributions, with the actions that find minimums of the energy function being the most likely actions given the expert dataset, which contain possibly conflicting navigation strategies.
%
These energy landscapes act like probability distributions, where low-energy actions are the most likely, even under conflicting expert strategies.
Thus, as seen in Fig.~\ref{fig:multimodal-energy}, both turning behaviors were learned and modeled by the EBM even though no explicit command of turning left or right were present in the expert training set.
Notably, the energy landscapes exhibit flatter minima in multimodal settings and sharper minima in unimodal ones—such as the third column in Fig.~\ref{fig:multimodal-energy-stop}, corresponding to stopping at a red light. Additionally, the ten DFO inferences tend to cluster tightly, suggesting that the model concentrates probability mass in small regions of the action space.

%%%%%%%%%%%%%%%%%%%%%%%%%%%%%%%%%%%%%%%%%%%%%%%%
The average traveled distances of the BC, IBC, and DA-IBC agents in Town2 were 9,074, 4,855, and 12,741, respectively. Our proposed DA-IBC significantly outperformed the baseline IBC in terms of traveled distance. However, it exhibited a higher traffic light infraction rate compared to BC—an issue that we aim to address in future work. We also hypothesize that the higher infraction rate observed in IBC may stem from its greater exploratory behavior. Unlike BC, which tends to mimic expert demonstrations conservatively, IBC explores a broader range of scenarios, potentially encountering more opportunities for infractions in the absence of explicit penalty mechanisms.
%
% The average traveled distances of BC, IBC and DA-IBC agents in \textit{town2} correspond to 9074, 4855, and 12741, respectively. Our proposed DA-IBC improved considerably over vanilla IBC, but presented a higher traffic light infraction rate than BC, an issue to be tackled in future work. We also argue that IBC, without a special penalization mechanism for infractions, might present a higher infraction rate because it explores more the environment, encountering more situations than the BC agent.

% \begin{table}[htbp]
% \centering
% \caption{Evaluation metrics on the CARLA driving benchmark.  
% Values are reported as \emph{mean} $\pm$ \emph{standard deviation}.  
% For \textbf{Infraction Penalty}, higher is better.}
% \label{tab:carla_metrics}
% \begin{tabular}{l|c|c|c}
% \toprule
% % \hline
% \textbf{Method} & \textbf{Distance (m)} & \textbf{Infraction Penalty} & \textbf{Score} \\
% \hline
% \midrule
% BC      & $9074 \,\pm\, 593$   & $0.97 \,\pm\, 0.09$ & $8807 \,\pm\, 1040$ \\
% IBC     & $4855 \,\pm\, 4706$  & $0.68 \,\pm\, 0.25$ & $2937 \,\pm\, 2976$ \\
% DA‑IBC  & $12741 \,\pm\, 209$  & $0.66 \,\pm\, 0.18$ & $8456 \,\pm\, 2322$ \\
% \bottomrule
% \end{tabular}
% \end{table}

\section{Conclusion}
In this work, we proposed an EBM architecture for autonomous vehicle navigation in urban environments simulated in CARLA, that maps an observation-action pair to an energy value. As the same observation (a Bird's-Eye View of the vehicle scene, last actions, and speed) can be paired to multiple conflicting actions in the expert training set,
%, i.e., the expert set may contain different driving styles, 
the EBM can model multimodal action distributions by learning an energy landscape as a function of the action input. 
% We proposed DA-IBC, Data-Augmented IBC, which: adapted the inference DFO algorithm for IBC, relying on the initialization of the actions that samples from the expert set instead of the uniform distribution; and updated the IBC training loss function by generating counterexamples also from the expert's actions with an added perturbation.
% We proposed DA-IBC (Data-Augmented IBC), which extends standard IBC in two key ways: first, by adapting the inference DFO algorithm to initialize action samples from the expert dataset rather than a uniform distribution; and second, by updating the training loss function to generate counterexamples specifically from the expert’s actions, with added perturbation.
We proposed DA-IBC (Data-Augmented IBC), an extension of standard IBC that introduces two key innovations: (1) it modifies the inference DFO algorithm to initialize action samples from the expert dataset instead of a uniform distribution; and (2) it updates the training loss to generate counterexamples by perturbing the expert’s actions,
refining the energy function to better separate expert behavior from nearby alternatives.

% We considered agents without the BEV's route channel in order to more easily test the multimodality situations in different T-intersections of a new test city \textit{town02}, with the agent trained in \textit{town01}. 
We omitted the BEV route channel to better test multimodal decisions at T-intersections in a new city (\textit{town02}), using agents trained in \textit{town01}.
In this way, even though the navigation is routeless, it serves the purpose to evaluate the capacity of the EBM to learn the multimodal distributions, which we have shown to 
be successful. 
% Furthermore, 
% even though the BC agent was able to drive equally well as IBC, 
% BC is not capable of learning multimodal behavior as IBC can.
Future works include the extension to cities with pedestrians and other vehicles, and the design of a goal-directed agent from a routeless model as trained in this work, which can take high-level commands in order to reach a destination.

\bibliographystyle{sbc}
\bibliography{biblio}

\end{document}